\documentclass{article}
\usepackage{ijcai20}

\usepackage{times}
\usepackage{soul}
\usepackage{url}
\usepackage[hidelinks]{hyperref}
\usepackage[utf8]{inputenc}
\usepackage[small]{caption}
\usepackage{graphicx}
\usepackage{amsthm}
\usepackage{booktabs}
\usepackage{algorithm}
\usepackage{algorithmic}
\usepackage{amsfonts} 
\usepackage{amsmath, bm}
\usepackage{multirow}
\usepackage{bbm}
\usepackage{color}
\newtheorem{theorem}{Theorem}
\newtheorem{corollary}{Corollary}

\title{Robust Fairness-aware Learning Under Sample Selection Bias}
\author{
Wei Du
\And
Xintao Wu 
\affiliations
University of Arkansas\\
\emails
\{wd005, xintaowu\}@uark.edu
}

\begin{document}
\maketitle

\begin{abstract}
The underlying assumption of many machine learning algorithms is that the training data and test data are drawn from the same distributions. However, the assumption is often violated in real world due to the sample selection bias between the training and test data. Previous research works focus on reweighing biased training data to match the test data and then building classification models on the reweighed training data. However, how to achieve fairness in the built classification models is under-explored. In this paper, we propose a framework for robust and fair learning under sample selection bias. Our framework adopts the reweighing estimation approach for bias correction and the minimax robust estimation approach for achieving robustness on prediction accuracy. Moreover, during the minimax optimization, the fairness is achieved under the worst case, which guarantees the model's fairness on test data.
We further develop two algorithms to handle sample selection bias when test data is both available and unavailable. 
We conduct experiments on two real-world datasets and the experimental results demonstrate its effectiveness in terms of both utility and fairness metrics.
\end{abstract}

\section{Introduction}
Traditional supervised machine learning assumes that training data and test data are independently and identically distributed (iid), i.e., each example $t$ with pairs of feature input $x$ and label $y$ drawn from the same distribution $\mathcal{Q} = P(x,y)$.  The conditional label distribution, $P(y|x)$, is estimated as $\hat{P}(y|x)$ (aka, a classifier $f(x)$) from the given training dataset $\mathcal{D}_s$. Similarly, in the fair machine learning, we aim to learn a fair classifier $f(x,a)$ from the training dataset drawn from $\mathcal{Q} = P(x,a,y)$ where $a$ is a protected attribute such as gender or race. However, when the distributions on training and test data sets do not match, we are facing sample selection bias or covariate shift. Sample selection bias occurs frequently in reality --- the available training data have been collected in a biased manner whereas the test is instead performed over a more general population. The classifier $f$ simply learned from the training dataset is vulnerable to sample selection bias and will incur more accuracy loss over test data. Moreover, the fair classifier trained with the biased data cannot guarantee fairness over test data. This is a serious concern when it is critical and imperative to achieve fairness in many applications.  

In this paper, we  develop a framework for robust and fair learning under sample selection bias. We embrace the uncertainty incurred by sample selection bias by producing predictions that are both fair and robust in test data. Our framework adopts the reweighing estimation approach for bias correction and the minimax robust estimation approach for achieving robustness on prediction accuracy. Moreover, during the minimax optimization, the fairness is achieved under the worst case, which guarantees the model's fairness on test data. To address the intractable issue, we approximate the fairness constraint using the boundary fairness and combine into the classifier's loss function as a penalty. The modified loss function is minimized in view of the most adverse distribution within a Wasserterin ball centered at the empirical distribution of the training data.

We present  two algorithms, \textit{RFLearn}\textsuperscript 1 for the scenario where the unlabeled test dataset $\mathcal{D}$  is available, and \textit{RFLearn}\textsuperscript 2 for the scenario where $\mathcal{D}$ is unavailable. In \textit{RFLearn}\textsuperscript 1,  we estimate the sample selection probability via its density ratio of training data and test data and then correct the bias in loss function. 

In \textit{RFLearn}\textsuperscript 2, we introduce some natural assumptions, i.e., the samples in the same cluster have the same  selection probability which is within a range from the uniform selection probability. The algorithm first clusters the training data and robustifies the sample selection probability estimation of each cluster within a Wasserterin ball. We test our algorithms on two real-world datasets and experimental results demonstrate that our algorithms can achieve both good performance on prediction and fairness.

\section{Problem Formulation}
We first define notations throughout this paper. Let $X$ denote the feature space, $A$ the protected attribute, and $Y$ the label set. 
Let $\mathcal{Q}$ denote the true distribution over $X \times A \times Y$ according to which test samples $t = (x,a,y)$ are drawn. 

For simplicity, we  assume both $y$  and $a$ are binary where $y = 1$ ($0$) denotes the favorable (unfavorable) decision and $a = 1$ ($0$) denotes the majority (minority) group.
Under the sample selection bias setting, the learning algorithm receives a training dataset $\mathcal{D}_s$ of  $N_{\mathcal{D}_s}$ labeled points  $t_1,\cdots,t_{N_{\mathcal{D}_s}}$ drawn according to a biased distribution $\mathcal{Q}_s$ over $X \times A \times Y$. This sample bias can be represented by a random binary variable $s$ that controls the selection of points, i.e., $s = 1$ for selected and $s=0$ otherwise.  
We consider the problem of  building over  $\mathcal{D}_{s}$  a fair classifier $f$ assigning labels $\hat{Y} \in \{0,1\}$ that depends only on $X,A$ (so $\hat{Y} \perp (Y,S)|X,A$) under certain fairness constraints. Many fairness notions  were proposed in the literature, such as demographic parity and mistreatment parity. In this paper, we choose the classic demographic parity and use  risk difference ($RD$) as the fairness quantity. In short, $RD$ measures the difference of the positive predictions between the majority group and minority group.

\noindent\textbf{Problem Formulation 1} (Fair Classifier Under Sample Selection Bias)
With the observed $\mathcal{D}_{s}$, how to construct a fair classifier $f$ that minimizes the expected loss $\mathbb{E}_{(x,a,y) \in \mathcal{Q}}[l(f(x, a), y)]$ subject to $|RD(\mathcal{Q})| \leq \tau$ where $l$ is the loss function,  $RD(\mathcal{Q})$ is the risk difference over distribution $\mathcal{Q}$, i.e., $RD(\mathcal{Q})=
|P_{\mathcal{Q}}(\hat{y}=1|a = 1)- P_{\mathcal{Q}}(\hat{y}=1|a = 0)|$, and $\tau \in [0, 1]$ is a threshold for the fairness constraint.

\section{Fair Classifier under Sample selection bias}

The probability of drawing $t=(x,a, y)$ according to the true but unobserved distribution $\mathcal{Q}$ is straightforwardly related to the observed distribution $\mathcal{Q}_s$. By definition of the random selection variable $s$, the observed biased distribution  $\mathcal{Q}_s$ can be expressed by $P_{\mathcal{Q}_s}(t) =  P_\mathcal{Q}(t|s = 1)$ or $P_{\mathcal{Q}_s}(x, a, y) =  P_\mathcal{Q}(x, a, y|s = 1)$. Assuming $P(s=1|x, a) \neq 0$ for all $t \in X \times A \times Y$, by the Bayes formula, we have
\begin{equation}
     P_\mathcal{Q}(t) = \frac{P(t|s=1)P(s=1)}{P(s=1|x, a)}= \frac{P(s=1)}{P(s=1|x, a)}P_{\mathcal{Q}_s}(t)
\end{equation}
Hence, if we define and  construct the new distribution $\hat{\mathcal{Q}}_s$ as $\frac{P(s = 1)}{P(s = 1|t)}{\mathcal{Q}_s}$, i.e., $P_{\hat{\mathcal{Q}}_s}(x, a, y) =  \frac{P(s = 1)}{P(s = 1|x, a,y)}P_{\mathcal{Q}_s}(x, a, y)$, we have the following result.
\begin{equation}
\begin{aligned}
&\mathbb{E}_{(x,a,y) \in \hat{\mathcal{Q}}_s}[l(f(x, a), y)] 
 = \sum_{x,a,y}l(f(x, a), y))P_{\hat{\mathcal{Q}}_s}(x,a,y) \\
&=\sum_{x,a,y}{{l(f(x,a), y))} \frac{P(s = 1)}{P(s = 1|x, a,y)} P_{{\mathcal{Q}_s}}(x, a,y)} \\
&=\sum_{x,a,y}{{l(f(x,a), y))} \frac{P(s = 1)}{P(s = 1|x, a,y)} P_{{\mathcal{Q}}}(x, a,y|s=1)} \\
&=\sum_{x,a,y}{{l(f(x,a), y))} \frac{P(s = 1)}{P(s = 1|x, a,y)}
\frac{P_{\mathcal{Q}(s = 1|x, a, y)}P_{\mathcal{Q}}(x, a, y)}{P_{\mathcal{Q}}(s=1)}}\\
&=\sum_{x,a,y}l(f(x, a), y))P_{\mathcal{Q}}(x, a, y)\\
&= \mathbb{E}_{(x,a,y) \in \mathcal{Q}}[l(f(x, a), y)] \nonumber
\end{aligned}
\end{equation}
Similarly we have $ RD(\hat{\mathcal{Q}}_s) = RD(\mathcal{Q})$. Equivalently, if we define and construct a modified training dataset $\hat{\mathcal{D}}_s$ by introducing a weight $\frac{P(s = 1)}{P(s = 1|x, a,y)}$ to each record $t\in \mathcal{D}_s$, we can approximate $\mathbb{E}_{(x,a,y) \in \hat{\mathcal{Q}}_s}[l(f(x, a), y)]$ using $\mathbb{E}_{(x,a,y) \in \hat{\mathcal{D}}_s}[l(f(x, a), y)]$, which can be expressed as $\frac{1}{N_{\mathcal{D}_s}}\sum_{i = 1}^{N_{\mathcal{D}_s}} \frac{P(s=1)}{P(s = 1|t)}l(f(x_i, a_i), y_i)$. 

If we can derive the probabilities $P(s=1)$ and $P(s=1|x, a)$, we could derive the true probability $P_\mathcal{Q}$ from the biased one $P_{\mathcal{Q}_s}$ exactly and correct the sample selection bias in fair classification.

\begin{theorem}
Under sample selection bias, the classifier $f$ that minimizes 
$\mathbb{E}_{(x,a,y,s) \in \hat{\mathcal{D}}_s}[l(f(x, a), y)]$ subject to $RD(\hat{\mathcal{D}}_s) \leq \tau$ is a fair classifier. 
\label{Theorem:fairclassifier}
\end{theorem}

Note that the classifier $f$ which tries to minimize $\frac{1}{N_{\mathcal{D}_s}}\sum_{i = 1}^{N_{\mathcal{D}_s}} l(f(x_i, a_i), y_i)$ under $RD(\mathcal{D}_s) \leq \tau$ would incur a large generalization error and cannot achieve fairness with respect to the true distribution $\mathcal{Q}$.

The sample selection bias causes training data to be selected non-uniformly from the population to be modeled. Generally there are four types of sample selection bias, missing completely at random when $P(s=1|x,a,y) = p(s=1)$, missing at random when $P(s=1|x,a,y) = p(s=1|x,a)$, missing at random-class when $P(s=1|x,a,y) = p(s=1|y)$, and missing not at random when there is no independence assumption between $x$, $a$, $y$ and $s$. 
In the following, we focus on missing at random, which indicates biasedness of the selected sample depends on the feature vector $x$ and protected attribute $a$.

The fairness constraint $RD(\hat{\mathcal{D}}_s) \leq \tau$ can be derived as 
\begin{equation}\label{eq:risk_difference_D_s_hat}
\resizebox{1.0\hsize}{!}{
    $|\frac{\sum \mathbbm{1}_{(x_i, a_i) \in \mathcal{D}_s^{11}}\frac{P(s=1)}{P(s=1|x_i, a_i)}}{\sum\mathbbm{1}_{(x_i, a_i) \in \mathcal{D}_s^{\cdot 1}}\frac{P(s=1)}{P(s=1|x_i, a_i)}} - \frac{\sum \mathbbm{1}_{(x_i, a_i) \in \mathcal{D}_s^{10}}\frac{P(s=1)}{P(s=1|x_i, a_i)}}{\sum\mathbbm{1}_{(x_i, a_i) \in \mathcal{D}_s^{\cdot 0}}\frac{P(s=1)}{P(s=1|x_i, a_i)}}| \leq \tau$},
\end{equation}
 where $\mathbbm{1}_{[.]}$ is an indicator function, $\mathcal{D}_s^{ij} = \{(x_i, a_i)|\hat{Y} = i, A = j\}$ where $i,j \in \{0, 1\}$,  $\mathcal{D}_s^{\cdot j} = \{(x_i, a_i)|A = j\}$ where $j \in \{0, 1\}$ and $\cdot$ represents $\{0, 1\}$. 

Then the minimization of the loss on $\hat{\mathcal{D}}_s$ subject to the fairness constraint is as:
\begin{equation}\label{eq:reweigh_loss_fair}
\begin{aligned}
&\min_{\textbf{w} \in \mathcal{W}} L(\textbf{w}) = \frac{1}{N_{\mathcal{D}_s}}\sum_{i = 1}^{N_{\mathcal{D}_s}} \frac{P(s=1)}{P(s=1|x_i, a_i)}l(f(x_i, a_i), y_i)\\
&\text{subject to} \hspace{1cm} RD(\hat{D}_s) \leq \tau
\end{aligned}
\end{equation}
where $\textbf{w}$ are the parameters of the classifier $f$ and $R(\hat{D}_s)$ is shown in Eq. \ref{eq:risk_difference_D_s_hat}. We then have our following corollary. 

\begin{corollary}
With the assumption that selection variable $s$ and label $y$ are independent given $x$, i.e., $P(s|x, a, y) = P(s|x, a))$, the classifier $f$ that minimizes $\mathbb{E}_{(x,a,y,s) \in \hat{\mathcal{D}}_s}[l(f(x, a), y)]$ subject to $RD(\hat{\mathcal{D}}_s) \leq \tau$ is a fair classifier under  sample selection bias  where  $\hat{\mathcal{D}}_s$ is constructed by weighting each sample
$t\in \mathcal{D}_s$ with
$\frac{P(s = 1)}{P(s = 1|x, a)}$. 
\end{corollary}

\section{Robust Fairness-aware Learning}
To obtain the optimal solution of Eq. \ref{eq:reweigh_loss_fair}, we need to derive the sample selection probability $P(s = 1|t)$. However, it is rather challenging to get the true $P(s = 1|t)$ practically because the selection mechanism is usually unknown. Instead, we need to estimate the sample selection probability and use the estimated probability $\hat{P}(s = 1|t)$ as the true $P(s = 1|t)$. 

To take the estimation error between $P(s = 1|t)$ and $\hat{P}(s = 1|t)$ into consideration, we adopt the approach of minimax robust minimization \cite{hu2018does,liu2014robust,wen2014robust} which advocates for the worst case of any unknown true sample selection probability. We make an assumption here that the true $P(s=1|x,a)$ is with the $\epsilon$ range of the estimated $\hat{P}(s=1|x,a)$. Therefore, any value of $P(s=1|x,a)$ in this $\epsilon$ range represents the possible real unknown distribution $\mathcal{Q}$. 
Following the standard robust optimization approaches and taking the estimation error into consideration, we reformulate Eq. \ref{eq:reweigh_loss_fair}:
\begin{equation}\label{eq:robust_reweigh_loss_fair_general}
\scalebox{0.85}{
$\begin{aligned}
&\min_{\textbf{w} \in \mathcal{W}} \hspace{0.1cm} \max_{{P}(s=1|x_i, a_i)} \hspace{0.1cm} L(\textbf{w}, {P}) = \frac{1}{N_{\mathcal{D}_s}}\sum_{i = 1}^{N_{\mathcal{D}_s}} \frac{P(s=1)}{{P}(s=1|x_i, a_i)}l(f(x_i, a_i), y_i)\\
&\text{subject to}  \hspace{0.1cm}|P(s=1|x_i, a_i)-\hat{P}(s=1|x_i, a_i)| \leq \epsilon\\
& \hspace{1.5cm}RD(\hat{D}_s) \leq \tau 
\end{aligned}$
}
\end{equation}

In fact, $P(s=1)$ is a constant and does not affect the problem formulation and optimization. 
The robust minimax optimization can be treated as an adversarial game by two players. One player selects ${P}(s=1|x_i, a_i)$ within the $\epsilon$ range of the estimated $\hat{P}(s=1|x_i, a_i)$
to maximize the loss of the objective, which can be seen as the worst case of $\mathcal{Q}$. As aforementioned, the selection probability is determined by the corresponding distribution $\mathcal{Q}$ and then different selection probability can represent different $\mathcal{Q}$. Another player minimizes the worst case loss to find the optimal $\mathbf{w}$. There are two advantages of robust minimax optimization of Eq. \ref{eq:robust_reweigh_loss_fair_general}. First, it takes the worst case induced by the estimation error into consideration, thus the obtained classifier $f$ is robust to any possible $\mathcal{Q}$ within the error range of the estimation. Second, during the minimax optimization, the fairness is achieved under  the worst case. Therefore, we can guarantee the fairness  for any possible $\mathcal{Q}$ within the error range of the estimation. 

The computation of $RD(\hat{D}_s)$ involves the indicator function, as shown in Eq. \ref{eq:risk_difference_D_s_hat}, which makes it computationally intractable to reach the optimal solution of Eq. \ref{eq:robust_reweigh_loss_fair_general}.
To address the intractable issue, we  approximate the fairness constraint using the boundary fairness \cite{zafar2017fairness} and it has been proved that decision boundary fairness is a concept of risk difference. We define $C(t, \textbf{w})$ be the covariance between the sensitive attribute and the signed distance from the non-sensitive attribute vector to the decision boundary. Then we can write the boundary fairness on $\mathcal{D}_s$ as:
\begin{equation}\label{eq:fairness_boundary_D}
   C_{\mathcal{D}_s}(t, \textbf{w}) = \frac{1}{N_{\mathcal{D}_s}}\sum_{i = 1}^{N_{\mathcal{D}_s}}(a_i - \bar{a})d_{\textbf{w}(\textbf{x}_i)},
\end{equation}
where $a_i$ is the sensitive attribute value of $t_i$, $\bar{a} = \frac{1}{N_{\mathcal{D}_s}}{\sum_{i=1}^{N_{\mathcal{D}_s}}a_i}$ 
is the mean value of the sensitive attribute and $d_{\textbf{w}(\textbf{x}_i)}$ is the distance to the decision boundary of the classifier $f$ and is formally defined as $d_{\textbf{w}(\textbf{x}_i)} = \textbf{w}^T\textbf{x}_i$. Similarly we will have the boundary fairness on $\hat{\mathcal{D}}_s$ as:
\begin{equation}\label{eq:fairness_boundary_D_s}
     C_{\hat{\mathcal{D}}_s}(t, \textbf{w}) = \frac{1}{N_{\mathcal{D}_s}}\sum_{i = 1}^{N_{\mathcal{D}_s}}(a_i - \bar{a})\frac{P(s=1)}{{P}(s=1|x_i, a_i)}d_{\textbf{w}(\textbf{x}_i)}.
\end{equation}
We enforce $C_{\hat{\mathcal{D}}_s}(t, \textbf{w}) \leq \sigma, \sigma \in R^+$ to achieve the fair classification. With the boundary fairness, we can rewrite the robust and fairness-aware loss function as:
\begin{equation}\label{eq:robust_reweigh_loss_fair_general_boundary_fairness}
\scalebox{0.85}{
$\begin{aligned}
&\min_{\textbf{w} \in \mathcal{W}} \hspace{0.1cm} \max_{{P}(s=1|x_i, a_i)} \hspace{0.1cm} L(\textbf{w}, {P}) = \frac{1}{N_{\mathcal{D}_s}}\sum_{i = 1}^{N_{\mathcal{D}_s}} \frac{P(s=1)}{{P}(s=1|x_i, a_i)}l(f(x_i, a_i), y_i)\\
&\text{subject to}  \hspace{0.1cm}|P(s=1|x_i, a_i)-\hat{P}(s=1|x_i, a_i)| \leq \epsilon\\
& \hspace{1.5cm}|\frac{1}{N_{\mathcal{D}_s}}\sum_{i = 1}^{N_{\mathcal{D}_s}}(a_i - \bar{a})\frac{P(s=1)}{{P}(s=1|x_i, a_i)}d_{\textbf{w}(\textbf{x}_i)}| \leq \sigma
\end{aligned}$}
\end{equation}

\subsection{Solving Robust Fairness-aware Optimization}
The optimization of Eq. \ref{eq:robust_reweigh_loss_fair_general_boundary_fairness} involves two sets of parameters $\textbf{w}$ and ${P}(s=1|x_i, a_i)$. According to its minimax formulation, it is preferable to obtain the optimal solution in an iterative manner by optimizing $\textbf{w}$ and ${P}(s=1|x_i, a_i)$ alternatively. 
First, we fix ${P}(s=1|x_1, a_i)$ and decompose the part of Eq. \ref{eq:robust_reweigh_loss_fair_general_boundary_fairness} only related to $\textbf{w}$ as:
\begin{equation}\label{eq:robust_reweigh_loss_fair_w}
\begin{aligned}
&\min_{\textbf{w} \in \mathcal{W}} \hspace{0.1cm} L(\textbf{w}) = \frac{1}{N_{\mathcal{D}_s}}\sum_{i = 1}^{N_{\mathcal{D}_s}} \frac{P(s=1)}{{P}(s=1|x_i, a_i)}l(f(x_i, a_i), y_i)\\
&\text{subject to}\hspace{0.5cm}|\frac{1}{N_{\mathcal{D}_s}}\sum_{i = 1}^{N_{\mathcal{D}_s}}(a_i - \bar{a})\frac{P(s=1)}{{P}(s=1|x_i, a_i)}d_{\textbf{w}(\textbf{x}_i)}| \leq \sigma
\end{aligned}
\end{equation}
It can be seen that with the fixed ${P}(s=1|x_i, a_i)$, the optimization of $\textbf{w}$ subjects to the linear fairness constraint. In fact, the optimal $\textbf{w}$ is determined by the choice of loss function $l$ and learning model $f$. 
For example, if the chosen model is regression model, then the optimization of $\textbf{w}$ subject to the linear constraints belongs to the family of quadratic programming, which can be solved efficiently by the available tools. Instead, if we choose some complex models, e.g. deep learning models, then the loss function parameterized by $\textbf{w}$ is non-convex, and it is quite challenging to apply the commonly used optimization techniques, e.g. gradient decent, to optimize $\textbf{w}$ with the existence of linear constraints. 
Therefore, for the generalization purpose, we choose to transform the fairness constraint as a penalty term and add to $L(\textbf{w})$, which can be expressed as:
\begin{equation}\label{eq:robust_reweigh_loss_fair_penalty_term}
\begin{aligned}
\min_{\textbf{w} \in \mathcal{W}} \hspace{0.1cm} L(\textbf{w}) = \frac{1}{N_{\mathcal{D}_s}}\sum_{i = 1}^{N_{\mathcal{D}_s}} \frac{P(s=1)}{{P}(s=1|x_i, a_i)}l(f(x_i, a_i), y_i)\\
+\beta(\frac{1}{N_{\mathcal{D}_s}}\sum_{i = 1}^{N_{\mathcal{D}_s}}(a_i - \bar{a})\frac{P(s=1)}{{P}(s=1|x_i, a_i)}d_{\textbf{w}(\textbf{x}_i)} -\sigma)^2
\end{aligned}
\end{equation}
where $\beta$ is a hyperparameter that controls the trade-off between the utility and fairness. By the transformation, standard optimization techniques such as stochastic gradient decent can be used to solve Eq. \ref{eq:robust_reweigh_loss_fair_penalty_term}.

Second, we can fix $\textbf{w}$ and decompose the part of Eq. \ref{eq:robust_reweigh_loss_fair_general_boundary_fairness} only related to ${P}(s=1|x_i, a_i)$ as the following:
\begin{equation}\label{eq:robust_reweigh_loss_fair_P_hat}
\scalebox{0.9}{
$\begin{aligned}
&\max_{{P}(s=1|x_i, a_i)} \hspace{0.1cm} L({P}) = \frac{1}{N_{\mathcal{D}_s}}\sum_{i = 1}^{N_{\mathcal{D}_s}} \frac{P(s=1)}{{P}(s=1|x_i, a_i)}l(f(x_i, a_i), y_i)\\
&\text{subject to}  \hspace{0.1cm}|P(s=1|x_i, a_i)-\hat{P}(s=1|x_i, a_i)| \leq \epsilon\\
& \hspace{1.5cm}|\frac{1}{N_{\mathcal{D}_s}}\sum_{i = 1}^{N_{\mathcal{D}_s}}(a_i - \bar{a})\frac{P(s=1)}{{P}(s=1|x_i, a_i)}d_{\textbf{w}(\textbf{x}_i)}| \leq \sigma
\end{aligned}$}
\end{equation}

The objective of Eq. \ref{eq:robust_reweigh_loss_fair_P_hat} is a linear combination of $\frac{P(s=1)}{{P}(s=1|x_i, a_i)}$ as we can treat $\frac{P(s=1)}{{P}(s=1|x_i, a_i)}$ to be one variable. $P(s=1)$ is a constant and does not affect the optimization.  For the first constraint, $|P(s=1|x_i, a_i)-\hat{P}(s=1|x_i, a_i)| \leq \epsilon$, we can obtain the range of $\frac{P(s=1)}{{P}(s=1|x_i, a_i)}$ after we estimate the range of each $P(s=1|x_i, a_i)$. The second constraint  $|\frac{1}{N_{\mathcal{D}_s}}\sum_{i = 1}^{N_{\mathcal{D}_s}}(a_i - \bar{a})\frac{P(s=1)}{{P}(s=1|x_i, a_i)}d_{\textbf{w}(\textbf{x}_i)}| \leq \sigma$ in Eq. \ref{eq:robust_reweigh_loss_fair_P_hat} is linear with respect to $\frac{P(s=1)}{{P}(s=1|x_i, a_i)}$ when $\textbf{w}$ is fixed.   Therefore, the optimization of $\frac{P(s=1)}{{P}(s=1|x_i, a_i)}$ is a standard linear programming and we can directly apply linear programming tool to get the optimal solution of $\frac{P(s=1)}{{P}(s=1|x_i, a_i)}$ without any additional relaxation.

\subsection{\textit{RFLearn}\textsuperscript{1}: Sample Bias Correction with $\mathcal{D}$}
In the previous section we formulate the robust and fairness-aware problem under the assumption that the estimated $\hat{P}(s=1|x, a)$ is obtained, but without addressing how to estimate it. In this section, we assume the unlabeled $\mathcal{D}$ is available and  estimate the true $P(s=1|x, a)$ by using $\mathcal{D}$ and $\mathcal{D}_s$  together. 
For a particular data record $t$, the ratio between the number of times $t$ in $\mathcal{D}$ and the number of times $t$  in $\mathcal{D}_s$ in terms of $(a,x)$ is an estimation value for $P(s=1|x, a)$. Formally, for $t \in \mathcal{D}$, let $\mathcal{D}^t$ denote the subset of $\mathcal{D}$ containing exactly all the instances of $t$ and $n_t = |\mathcal{D}^t|$. 
Similarly, let $\mathcal{D}_s^t$ denote the subset of $\mathcal{D}_s$ containing exactly all the instances of $t$ and $m_t = |\mathcal{D}_s^t|$. We then have $\hat{P}(s=1|x, a) = \frac{m_t}{n_t}$.

\vspace{0.2cm}
\noindent\textbf{Lemma 1} \cite{cortes2008sample}
Let $\delta > 0$, then, with probability at least $1 - \delta$, the following inequality holds for all $t \in \mathcal{D}_s$:
\begin{equation}\label{eq:upper_theory}
    |P(s=1|x, a) - \frac{m_t}{n_t}| \leq \sqrt{\frac{\text{ln}2m' + \text{ln}\frac{1}{\delta}}{p_0N_{\mathcal{D}}}}
\end{equation}
where  $m'$ denotes the number of unique points in $\mathcal{D}_s$ and $p_0= \text{min}_{t \in \mathcal{D}}P(t) \neq 0$. 

For notation convenience, we define $\theta(x_i, a_i) = P(s=1|x_i, a_i)$ and $\hat{\theta}(x_i, a_i) = \hat{P}(s=1|x_i, a_i)$, where $\hat{\theta}(x_i, a_i)$ is the empirical value based on the frequency estimation. 
Lemma 1 states that $|\theta(x_i, a_i) - \hat{\theta}(x_i, a_i)|$ is upper bounded by the right term in Eq. \ref{eq:upper_theory}. Then we can apply the robust fairness aware framework by setting $\theta(x_i, a_i)$ within $\epsilon$ range of estimated $\hat{\theta}(x_i, a_i)$, where $\epsilon$ is the right term of Eq. \ref{eq:upper_theory}. According to the Theorem 2 in \cite{cortes2008sample}, the generalization error between the true distribution $\theta(x_i, a_i)$ and distribution using the estimated $\hat{\theta}(x_i, a_i)$ is expressed as:
\begin{equation}\label{eq:generalization_error}
    |L_{\theta}(\textbf{w}) - L_{\hat{\theta}}(\textbf{w})| < \mu \sqrt{\frac{\text{ln}2m' + \text{ln}\frac{1}{\delta}}{p_0N_{\mathcal{D}}}}
\end{equation}
where $\mu$ is a constant determined by $\sigma$ (Lemma 1) and hyperparameter $\beta$ (Eq. \ref{eq:robust_reweigh_loss_fair_penalty_term}). Suppose the maximum value of $L_{\hat{\theta}}(\textbf{w})$ is defined as $L_{\hat{\theta}}(\textbf{w})_{max}$ and our robust fairness-aware optimization is to minimize $L_{\hat{\theta}}(\textbf{w})_{max}$ per iteration. The loss $L_{\hat{\theta}}(\textbf{w})_{max}$ consists of both the prediction loss and fairness loss. Therefore, the minimization of upper bound of the generalization error of true distribution can provide robustness in terms of both prediction and fairness.

\begin{table*}
  \centering
  \caption{Model performance under data distribution shift (Adult and Dutch) Acc: accuracy}
  \scalebox{0.9}{
  \begin{tabular}{|c|c|c|c|c|c|c|c|c|}
  \hline
	\multirow{2}*{\textbf{Methods}}  & \multicolumn{4}{|c|}{\textbf{Adult Dataset}} & \multicolumn{4}{|c|}{\textbf{Dutch Dataset} }\\
	\cline{2-9}
     & {Training Acc} & {Test Acc} & {Training $RD$} & {Test $RD$} & {Training Acc} & {Test Acc} & {Training $RD$} & {Test $RD$} \\
    \hline
    \hline
     \textit{LR (unbiased)}  &  0.8124 &  0.8126 & 0.1562 & 0.1373 & 0.7493 &  0.7669 & 0.1498 & 0.1395  \\
     \hline
     \textit{LR}  &  0.8041 &  0.7882 & 0.1344 & 0.1228 & 0.7018 & 0.6821  & 0.0378 & 0.1012 \\
     \hline
     \textit{FairLR}  &  0.7823 &  0.7622 & 0.0231 & 0.0956 & 0.7006 &  0.6624 & 0.0289 & 0.0991 \\
     \hline
      \textit{\cite{wen2014robust}}  &  0.7883 &  0.8048 & 0.1348 & 0.1333 & 0.6812 & 0.7044  & 0.1421 & 0.1394 \\
     \hline
     \hline
     $\textit{RFLearn}^{1-}$ & 0.7412 &  0.7875 & 0.0351 & 0.1048 & 0.6501 & 0.6879 & 0.0315 & 0.0809 \\
     \hline
     $\textit{RFLearn}^1$ & 0.7484 &  0.7816 &  0.0281 & 0.0416 & 0.6673 & 0.6910 & 0.0317 & 0.0405 \\
     \hline
     \hline
     $\textit{RFLearn}^{2-}$ &  0.7473 &  0.7771 & 0.0321 &  0.0963 & 0.6457 & 0.6809  & 0.0411 & 0.0973 \\
     \hline
     $\textit{RFLearn}^2$ &  0.7336 &  0.7678 &  0.0197 & 0.0238 & 0.6479 & 0.6755  & 0.0321 & 0.0373 \\
     \hline
  \end{tabular}}
  \label{tab: accuracy_data_shift}
\end{table*}

\begin{table*}
  \centering
  \caption{Model performance of $\textit{RFLearn}^1$ under sample selection bias with different $\delta$ (Adult and Dutch). Acc: accuracy}
  \scalebox{0.9}{
  \begin{tabular}{|c|c|c|c|c|c|c|c|c|}
  \hline
	\multirow{2}*{$\delta$} & \multicolumn{4}{|c|}{\textbf{Adult Dataset}} & \multicolumn{4}{|c|}{\textbf{Dutch Dataset} }\\
	\cline{2-9}
     & {Training Acc} & {Test Acc} & {Training $RD$} & {Test $RD$} & {Training Acc} & {Test Acc} & {Training $RD$} & {Test $RD$} \\
     \hline
     0.025  &  0.7181 &  0.7601 & 0.0189 &  0.0219 & 0.6521 & 0.6812  & 0.0291& 0.0326\\
    \hline
     0.05  &  0.7217 &  0.7673 & 0.0239 &  0.0398 & 0.6521 & 0.6812 & 0.0321 & 0.0326\\
     \hline
     0.1 & 0.7484 &  0.7816 &  0.0307 & 0.0416 & 0.6673 &  0.6910 & 0.0378 & 0.0405 \\
     \hline
    0.15 & 0.7239 &  0.7768 & 0.0277 & 0.0333 & 0.6701 & 0.6994  & 0.0275 & 0.0379 \\
     \hline
  \end{tabular}}
  \label{tab: accuracy_delta}
\end{table*}

\begin{table*}
  \centering
  \caption{Model performance of $\textit{RFLearn}^2$ under sample selection bias with different $\rho$ (Adult and Dutch). Acc: accuracy}
  \scalebox{0.9}{
  \begin{tabular}{|c|c|c|c|c|c|c|c|c|}
  \hline
	\multirow{2}*{$\rho$} & \multicolumn{4}{|c|}{\textbf{Adult Dataset}} & \multicolumn{4}{|c|}{\textbf{Dutch Dataset} }\\
	\cline{2-9}
     & {Training Acc} & {Test Acc} & {Training $RD$} & {Test $RD$} & {Training Acc} & {Test Acc} & {Training $RD$} & {Test $RD$} \\
    \hline
     0.2  &  0.7229 &  0.7558 & 0.0178 &  0.0114 & 0.6401 & 0.6543  &0.0175 & 0.0214\\
     \hline
     0.4 & 0.7336 &  0.7628 & 0.0197 & 0.0238 & 0.6479 & 0.6755 &0.0321 & 0.0373 \\
     \hline
     0.6 & 0.7428 &  0.7724 &  0.0269 & 0.0361 & 0.6544 &  0.6792 & 0.0301 & 0.0314 \\
     \hline
  \end{tabular}}
  \label{tab: accuracy_rho}
\end{table*}

\subsection{\textit{RFLearn}\textsuperscript{2}: Sample Bias Correction without $\mathcal{D}$}

In this section, we focus on the scenario without unlabeled test data $\mathcal{D}$. The challenge is how to use $\mathcal{D}_s$ alone to estimate the true sample selection probability so that we can construct $\hat{\mathcal{D}}_s$ to resemble $\mathcal{D}$. In general, the exact relationship between $\mathcal{D}_s$ and $\mathcal{D}$ is unknown. Without additional assumptions, it is impossible to build a model to resemble the true $\mathcal{D}$ with only access to the observed $\mathcal{D}_s$.

We assume that 1)  there exist $K$ clusters in $\mathcal{D}_s$; 2) the samples in the same cluster have the same selection probability; and 3) the selection probability of each sample is within a range of the uniform selection probability. Under these assumptions,   the ratio $\frac{P(s=1)}{{P}(s=1|x_i, a_i)}$ for each sample from the same cluster is the same. The ratio vector for $K$ clusters is defined as $\mathbf{r} = (r_1, r_2, \cdots, r_K)$. We  robustify the estimation by approximating $r$ within a Wasserstein ball $B_{\rho}$ \cite{blanchet2019robust} around the uniform ratio $\mathbf{r}_u$, where all of the values in $\mathbf{r}_u$ is 1. Formally we have $|\mathbf{r} - \mathbf{r}_u| \leq \rho$, where $\rho$ is the radius of the Wasserstein ball. 
Suppose $x_i$ belongs to the $k$-th cluster where $ k \in [K]$, we can write the robust loss of $f$ with only $\mathcal{D}_s$ available as the following:
\begin{equation}\label{eq:robust_reweigh_loss_unknown_D_cluster}
\begin{aligned}
&\min_{\textbf{w} \in \mathcal{W}} \hspace{0.1cm} \max_{\mathbf{r} \in B_{\rho}} \hspace{0.1cm} L(\textbf{w},\mathbf{r}) = \frac{1}{N_{\mathcal{D}_s}}\sum_{i = 1}^{N_{\mathcal{D}_s}} r_k l(f(x_i, a_i), y_i)\\
&\text{subject to}  \hspace{1cm} |\mathbf{r} - \mathbf{r}_u| \leq \rho \\
&\hspace{2.3cm}|\frac{1}{N_{\mathcal{D}_s}}\sum_{i = 1}^{N_{\mathcal{D}_s}}(a_i - \bar{a})r_k d_{\textbf{w}(\textbf{x}_i)}| \leq \sigma
\end{aligned}
\end{equation}
The above formulation also saves huge amounts of computational cost as it  reduces the number of constraints from the training data size level to the cluster size level.

\section{Experiment}
\textbf{Dataset.} We use two benchmark datasets, Adult  \cite{Dua:2019} and Dutch  \cite{vzliobaite2011handling}, to  evaluate our proposed algorithms, \textit{RFLearn}\textsuperscript 1 and \textit{RFLearn}\textsuperscript 2. Adult dataset consists of individual's information such as age, education level, gender, occupation, race, and so forth. The task for Adult dataset is to predict whether an individual's income is over 50k or not based on the collected personal information. Dutch dataset is a collection of Census data in Netherlands, where each data sample has different attributes like Adult datasets, including education level, gender, race, marital status and so forth. The task is also a binary classification to predict whether an individual belongs to low income or high income group. 
In both datasets, we set \textit{gender} as the protected attribute and  use one-hot encoding to convert the categorical attributes to vectors and apply normalization to covert numerical attributes into the range $[0, 1]$. After preprocessing, Adult has 45222 records and 40 features, while Dutch has 60420 records and 35 features.

\noindent\textbf{Experimental Setting}. 
We follow the biased data generation approach in \cite{laforgue2019statistical} and choose to select the data based on the education level (married status) for Adult (Dutch). For Adult, we create the biased training data by selecting 18157 records from the first 35000 records (randomly choosing 70\% of people with 10+ years education and 30\% of people with less than 10 years education) and use the rest 10222 records as the test data.  Similarly for Dutch, we create the biased training data by selecting 20928 examples from the first 45000 records (randomly choosing 75\% of married people and 25\% of unmarried people) and use the rest 15420 records as the test data. 

\noindent \textbf{Baselines.} We choose the logistic regression (LR) to implement and evaluate different algorithms. We consider the following baselines to compare with our proposed \textit{RFLearn}\textsuperscript 1 and \textit{RFLearn}\textsuperscript 2: (a) LR without fairness constraint (\textit{LR}); (b) LR with fairness constraint (\textit{FairLR}); (c) robust LR in \cite{wen2014robust} that uses kernel functions to reweigh samples under covariate shift but ignores the fairness constraint.
For \textit{RFLearn}\textsuperscript {1} (\textit{RFLearn}\textsuperscript {2}), we also consider its variation  \textit{RFLearn}\textsuperscript {1-} (\textit{RFLearn}\textsuperscript {2-}) that optimizes the robust loss with unweighted fairness constraint. 

\noindent \textbf{Metrics and Hyperparameters.} We evaluate the performance in terms of prediction accuracy and risk difference ($RD$). We consider a classifier is fair if $RD(f) \leq 0.05$. The hyperparameter $\beta$ that controls the accuracy-fairness trade-off is set as 1. The radius $\rho$ is set as 0.4, $\delta$ is set as 0.1, and $\sigma$ is set as 0.2. 
For $\textit{RFLearn}^{2-}$ and $\textit{RFLearn}^{2}$, we apply K-means \cite{hartigan1979algorithm} to cluster the training data and set the number of clusters 300. We run all experiments 20 times and report the average results.

\subsection{Accuracy vs. Fairness}
We  report the accuracy and fairness on both training data and test data for each method in our results. Note that our goal is to achieve robust accuracy and fairness on test data under the sample selection bias. Results on both training and test data help explain why baselines focusing on the training data itself fail achieving good accuracy and fairness on test data.

\noindent \textbf{Accuracy.} Table \ref{tab: accuracy_data_shift} shows prediction accuracy and risk difference of each model on training and test data for both Adult and Dutch. The first row \textit{LR (unbiased)}  indicates the performance of the logistic regression model with unbiased training data. We randomly select the same number of data from the original datasets.  The results of the rest rows are all conducted on our biased training data discussed in the experimental setting section. We can draw several important points from the experimental results. 

First, the testing accuracy of \textit{LR} is 0.7882 for Adult and 0.6821 for Dutch, whereas the accuracy of \textit{LR (unbiased)} is 0.8126 for Adult and 0.7669 for Dutch. It demonstrates that the model prediction performance degrades under sample selection bias. Second, with the use of robust learning, the prediction accuracy for Adult (Ducth) of \textit{RFLearn}\textsuperscript {1-} is 0.7875 (0.6879) and \textit{RFLearn}\textsuperscript {2-} is 0.7771 (0.6809), which outperforms \textit{FairLR} by 0.0253 (0.0255) and 0.0109 (0.0185). As the fairness constraints of \textit{RFLearn}\textsuperscript {1-} and \textit{RFLearn}\textsuperscript {2-} are the same with \textit{FairLR}, it demonstrates that the robust learning in Eq. \ref{eq:robust_reweigh_loss_fair_general_boundary_fairness} can provide robust prediction under the sample selection bias. Third, the testing accuracy from robust learning methods is higher than the training accuracy, which further demonstrates the advantage of robust learning under the sample selection bias. 
Fourth, the accuracy of \textit{RFLearn}\textsuperscript {1} is higher than that of \textit{RFLearn}\textsuperscript {2}. It is reasonable as we leverage the unlabeled test data in our \textit{RFLearn}\textsuperscript {1}.    

\noindent \textbf{Fairness.} In this section, we focus on fairness comparison. 
We summarize below the interesting observations from Table  \ref{tab: accuracy_data_shift}. First,  all of \textit{FairLR}, \textit{RFLearn}\textsuperscript {1-} and \textit{RFLearn}\textsuperscript {2-} can only achieve fairness on the training data with $RD \leq 0.05$, but none of these approaches can guarantee the fairness on the test data. 
\textit{RFLearn}\textsuperscript  {1-} and \textit{RFLearn}\textsuperscript {2-}, which uses the unweighted fairness constraint,  also  fail to achieve fairness on the test data. The method proposed by \cite{wen2014robust} only considers 
the robustness of prediction error but ignores the fairness, so that it cannot achieve the fairness on the test data as well. Second, \textit{RFLearn}\textsuperscript 1 and \textit{RFLearn}\textsuperscript 2 can achieve fairness on both training and test data. The result is consistent with our algorithms as we enforce the fairness under any possible adverse distribution. Therefore, the classifier is also robust in terms of meeting fairness requirement on the test data even it is trained on the biased sampling data.  

\subsection{Effects of the Hyperparameters}
In this section, we conduct sensitivity analysis of our \textit{RFLearn}\textsuperscript 1 and \textit{RFLearn}\textsuperscript 2
 with different hyperparameters. Due to space limits, we focus on two key parameters $\delta$ and $\rho$. Table \ref{tab: accuracy_delta} shows the performance of  \textit{RFLearn}\textsuperscript 1 with different $\delta$ values. Note that  in Lemma 1 the estimation error of the sample selection probability is upper bounded with  the probability greater than $1 - \delta$.  The upper bound (the right side in Eq. \ref{eq:upper_theory}) increases with the decreasing $\delta$. 
 A larger upper bound indicates that the adversary can generate more possible distributions during the robust optimization, hence
 helping achieve better prediction accuracy on test data. However, when the upper bound is too large,  excessive possible distributions may reduce the prediction accuracy on test data. The experimental results in Table \ref{tab: accuracy_delta} match our above analysis.  

Table \ref{tab: accuracy_rho} shows the result of \textit{RFLearn}\textsuperscript 2 under different radius $\rho$.  We can see that the testing accuracy increases with the increasing $\rho$. Larger $\rho$ indicates more possible generated distributions which are more likely to cover the test distribution and improve the model performance. Moreover, the proposed \textit{RFLearn}\textsuperscript 1 and \textit{RFLearn}\textsuperscript 2 can achieve both fairness on the training and test data with different $\delta$ and $\rho$. It is beneficial as the fairness performance of our proposed algorithms are not sensitive to the varying $\delta$ and $\rho$.

\section{Related Work}

There has been extensive research on classification under sample selection bias, e.g., \cite{DBLP:conf/icml/Zadrozny04,DBLP:conf/nips/HuangSGBS06}. \cite{laforgue2019statistical} provide a  methodology to address selection bias issues in statistical learning and extend the paradigmatic ERM approach to the sample selection bias setting.  Robust classification under sample selection bias has also been studied recently in \cite{hu2018does,liu2014robust,wen2014robust}. \cite{wen2014robust} consider covariate shift between the training and test data and apply Gaussian kernel functions to reweigh the training examples and correct 
the shift. They propose a minimax robust framework to minimize the most adverse distribution and provide robust classification when it comes to the test data. Similarly, \cite{liu2014robust} propose a robust bias-aware probabilistic classifier that can deal with different test data distributions using a minimax estimation formulation. 

Most recently, there have been a few papers that study fairness from the distributionally robust perspective.  \cite{taskesen2020distributionally} propose a distributionally robust logistic regression model with an unfairness penalty. They assume the unknown true test distribution is contained in a Wasserstein ball centered at the  empirical distribution on the observed training data. The proposed model, which robustifies the fair logistic regression against all distributions in the ball, is equivalent to a tractable convex problem when unfairness is quantified under the log-probabilistic equalized opportunities criterion.  However the approach robustifies the distribution at the individual data level and overlooks the overall distribution.
\cite{DBLP:conf/aaai/RezaeiFMZ20} propose the use of ambiguity set to derive the fair classifier based on the principles of distributional robustness. The proposed approach incorporates fairness criteria into a worst case logarithmic loss minimization but  ignores the distribution shift.  \cite{DBLP:conf/iclr/YurochkinBS20} develop an individual fair distributionally robust classifier with a Wasserstein ambiguity set. However, the approach does not admit a tractable convex reformulation. 

There has been research on fairness aware domain adaptation and transfer learning of fairness metrics, e.g., \cite{schumann2019transfer,coston2019fair,lahoti2020fairness,kallus2018residual}. In particular, \cite{schumann2019transfer} study the fair transfer learning from source domain to target domain and provide a fairness bound on the target domain of the predictor trained on the source domain data. \cite{coston2019fair} study the fair transfer learning with missing protected attributes under the covariate shift. 
Similarly, \cite{kallus2018residual} use covariate shift correction when computing fairness metrics to address bias in label collection. The above works also adopt the idea of reweighing the source examples to resemble the target domain examples during the training. However, they do not address the robustness in learning under sample selection bias. 

Related but different from our work is fair learning from label biased training data  \cite{DBLP:conf/forc/BlumS20,jiang2020identifying} and 
fair representation learning \cite{madras2018learning,song2019learning,zhao2019inherent}. For example,  \cite{jiang2020identifying} develop a  framework to model how label bias can arise in a dataset, assuming that there exists an unbiased ground truth,  and develop a bias correcting approach based on  re-weighting the training examples.  \cite{du2021fairness}  propose a robust framework to achieve fairness  in federated learning under covariate shift between the training data and test data. It applies Gaussian kernel to generate the adversary distribution and minimizes the worst adversary distribution loss during training. \cite{zhao2019inherent} investigate the relationship between fair representation and decision making with theoretical guarantee. They employ the adversarial technique to extract the fair representation. \cite{singh2021fairness} study the covariate shift in the domain adaption and assumes a known causal graph of the data generating process and a context variable causing the shift.  \cite{rezaei2020robust} apply minimax optimization to achieve robustness fairness under covariate shift, where the fairness violation penalty term between the target input distribution and adversary’s conditional label distribution is incorporated during the optimization. Our work is different from above works as we go from the sample selection probability estimation and minimize the worst case loss while enforcing fairness on re-weighted samples.

\section{Conclusion}
We have proposed a robust and fair learning framework to deal with the sample selection bias. Our framework adopts the reweighing estimation approach for bias correction and the minimax robust estimation  for achieving robustness on prediction accuracy and fairness on test data. We further developed two algorithms to handle sample selection bias when test data is both available and unavailable. 

Experimental results showed our algorithms can achieve both good prediction accuracy and fairness on test data. In our future work, we will investigate the sample selection bias under missing not at random, i.e., the sample selection probability also depends on the label. We will also study how to enforce other fairness notions such as equal opportunity\cite{hardt2016equality} .

\section{Acknowledgement}
This work was supported in part by NSF 1920920, 1940093, and 1946391,


\end{document}